\documentclass{article}
\usepackage{ijcai21}

\usepackage{times}

\usepackage{soul}
\usepackage{url}
\usepackage[hidelinks]{hyperref}
\usepackage[utf8]{inputenc}
\usepackage[small]{caption}
\usepackage{graphicx}
\usepackage{amsmath}
\usepackage{booktabs}
\title{Unsupervised Continual Learning via Self-Adaptive Deep Clustering Approach}

\author{
Mahardhika Pratama$^1$\footnote{Contact Author}\and
Andri Ashfahani$^1$\and
Edwin Lughofer$^{2}$
\affiliations
$^1$SCSE, Nanyang Technological University, Singapore\\
$^2$DKBMS, Johanes Kepler University, Linz, Austria \\
\emails
mpratama@ntu.edu.sg,
andriash001@e.ntu.edu.sg,
edwin.lughofer@jku.at
}

\begin{document}

\maketitle

\begin{abstract}
Unsupervised continual learning remains a relatively uncharted territory in the existing literature because the vast majority of existing works call for unlimited access of ground truth incurring expensive labelling cost. Another issue lies in the problem of task boundaries and task IDs which must be known for model's updates or model's predictions hindering feasibility for real-time deployment. Knowledge Retention in Self-Adaptive Deep Continual Learner, (KIERA), is proposed in this paper. KIERA is developed from the notion of flexible deep clustering approach possessing an elastic network structure to cope with changing environments in the timely manner. The centroid-based experience replay is put forward to overcome the catastrophic forgetting problem. KIERA does not exploit any labelled samples for model updates while featuring a task-agnostic merit. The advantage of KIERA has been numerically validated in popular continual learning problems where it shows highly competitive performance compared to state-of-the art approaches. Our implementation is available in \textit{\url{https://github.com/ContinualAL/KIERA}}.          
\end{abstract}

\section{Introduction}
	Continual learning is a machine learning field studying a learning model which can handle a sequence of tasks $T_1,T_2,T_3,...,T_K$ \cite{parisi2018continual} where $K$ labels the number of tasks. Unlike conventional learning problem with the i.i.d assumption, every task $T_k$ suffers from non-stationary conditions where there exists drifting data distributions $P(X,Y)_{k}\neq P(X,Y)_{k+1}$, emergence of new classes or combination between both conditions. The goal of general continual learning should be to build a model $f(.)$ which is capable of identifying and adapting to changes without suffering from the catastrophic forgetting problem. It is done on the fly with the absence of samples from previous tasks $T_{k-1}$. 
	
	There has been growing research interest in the continual learning domain in which the main goal is to resolve the issue of catastrophic forgetting thereby actualizing knowledge retention property of a model. These works can be divided into three categories: memory-based approach, regularization-based approach and structure-based approach. 
	\textbf{Memory-based Approach} is designed to handle the continual learning problem with the use of external memory storing past samples. Past samples are replayed along with new samples of the current task such that the catastrophic forgetting problem can be addressed. \textbf{Regularization-based approach} is put forward with the use of additional regularization term in the cost function aiming at achieving tradeoff points between old task and new task. \textbf{Structure-based approach} is developed under the roof of dynamic structure. It adds new network components to cope with new tasks while isolating old parameters from new task to avoid the catastrophic forgetting problem.
	
	The area of continual learning still deserves in-depth study because the vast majority of existing approaches are fully-supervised algorithms limiting their applications in the scarcity of labelled samples. Existing approaches rely on a very strong assumption of task boundaries and task IDs. This assumption is impractical because the point of changes is often unknown and the presence of task IDs for inference imply extra domain knowledge. 
	
	An unsupervised continual learning algorithm, namely Knowledge Retention in Self-Adaptive Deep Continual Learner (KIERA), is proposed in this paper. KIERA does not utilize any labelled samples for model updates. That is, labelled samples are only offered to establish clusters-to-classes associations required to perform classification tasks. KIERA is constructed from the idea of self-evolving deep clustering network making use of a different-depth network structure. Every hidden layer generates its own set of clusters and produces its own local outputs. This strategy enables an independent self-evolving clustering mechanism to be performed in different levels of deep embedding spaces. KIERA features an elastic network structure in which its hidden nodes, layers and clusters are self-generated from data streams in respect to varying data distributions.
	
	The parameter learning mechanism of KIERA is devised to achieve a clustering-friendly latent space via simultaneous feature learning and clustering. It puts forward reconstruction loss and clustering loss minimized simultaneously in the framework of stacked autoencoder (SAE) to avoid trivial solutions. The clustering loss is formed as the K-Means loss \cite{clustering_friendly} inducing the clustering-friendly latent space where data samples are forced to be adjacent to the winning cluster, i.e., the most neighboring cluster to a data sample. The centroid-based experience replay strategy is put forward to address the catastrophic interference problem. That is, the sample selection mechanism is carried out in respect to focal-points, i.e., data samples triggering addition of clusters. The selective sampling mechanism is integrated to assure a bounded replay buffer. The advantage of KIERA is confirmed with rigorous numerical study in popular problems where it outperforms prominent algorithms. 
	
	This paper presents four major contributions: 1) KIERA is proposed to handle unsupervised continual learning problems; 2) the flexible deep clustering approach is put forward in which hidden layers, nodes and clusters are self-evolved on the fly; 3) the different-depth network structure is designed making possible an independent clustering module to be carried out in different levels of deep embedding space; 4) the centroid-based experience replay method is put forward to address the catastrophic interference problem. 
	
\section{Related Works}
\noindent\textbf{Memory-based Approach}: iCaRL \cite{Rebuffi2017iCaRLIC} exemplifies the memory-based approach in the continual learning using the exemplar set of each class. The classification decision is calculated from the similarity degree of a new sample to the exemplar set. Another example of the memory-based approach is Gradient Episodic Memory (GEM) \cite{gem2017paz} where past samples are stored to calculate the forgetting case. A forgetting case can be examined from the angle between the gradient vector and the proposed update. This work is extended in \cite{Chaudy2019AGEM} and called Averaged GEM (AGEM). Its contribution lies in the modification of the loss function to expedite the model updates. Deep Generative Replay (DGR) \cite{shin2017continual} does not utilize external memory to address the catastrophic forgetting problem rather makes use of generative adversarial network (GAN) to create representation of old tasks. The catastrophic forgetting is addressed by generating pseudo samples for experience replay mechanism. Our work is framed under this category, because it can be executed without the presence of task IDs or task's boundaries. 
\newline\noindent\textbf{Regularization-based Approach}: Elastic Weight Consolidation (EWC) \cite{kirkpatrick2016overcoming} is a prominent regularization-based approach using the L2-like regularization approach constraining the movement of important network parameters. It constructs Fisher Information Matrix (FIM) to signify the importance of network parameters. Synaptic Intelligence (SI) \cite{zenke2017continual} offers an alternative approach where it utilizes accumulated gradients to quantify importance of network importance instead of FIM incurring prohibitive computational burdens. Memory Aware Synapses (MAS) \cite{Aljundi2018MemoryAS} is an EWC-like approach with modification of parameter importance matrix using an unsupervised and online criterion. EWC has been extended in \cite{schwarz2018progress} named onlineEWC where Laplace approximation is put forward to construct the parameter importance matrix. Learning Without Forgetting (LWF) is proposed in \cite{li2016learning} where it formulates a joint optimization problem between the current loss function and the knowledge distillation loss \cite{HinVin15Distilling}. This neuron-based regularization notion is presented in \cite{neuron_level} where the regularization is performed by adjusting the learning rates of stochastic gradient descent. In \cite{ISYANA}, the inter-task synaptic mapping is proposed. Notwithstanding that the regularization-based approach is computationally efficient, this approach requires the task boundaries and task IDs to be known.  
\newline\noindent\textbf{Structure-based Approach}: the catastrophic forgetting is overcome in progressive neural networks (PNNs) \cite{Rusu2016ProgressiveNN} by introduction of new column for every new task while freezing old network parameters. This approach is, however, not scalable for large-scale problem since the structural complexity linearly grows as the number of tasks. This drawback is addressed in \cite{Lee2018LifelongLW} with the use of selective retraining approach. Learn-to-grow is proposed in \cite{learntogrow} where it utilizes the neural architecture search to find the best network structure of a given task. The structure-based approach imposes expensive complexities.
\section{Problem Formulation}
The continual learning problem aims to build a predictive model $f(.)$ handling streaming tasks $T_1,T_2,T_3,...,T_K$ where $K$ denotes the number of tasks. Unlike conventional problems assuming the i.i.d condition, each task $T_k$ is influenced by non-stationary conditions. There exist two types of changes in the continual learning where the first one is known as the problem of changing data distributions while the second one is understood as the problem of changing target classes. The problem of changing data distributions or the concept drift problem is defined as a change of joint probability distribution $P(X,Y)_k\neq P(X,Y)_{k+1}$. The problem of changing target classes refers to different target classes of each task. Suppose that $L_k$ and $L_{k'}$ stand for the label sets of the $k-th$ task and the $k'-th$ task, $\forall k,k' L_{k}\cap L_{k'}=\emptyset$ for $k\neq k'$. This problem is also known as the incremental class problem. Each task normally consists of paired data samples $T_k=\{{x_n,y_n}\}_{n=1}^{N_k}$ where $N_k$ denotes the size of the $k-th$ task. $x_n\in\Re^{u\times u}$ is the $n-th$ input image while $y_n\in\{l_1,l_2,...,l_m\}$ is the target vector formed as a one-hot vector. This assumption, however, does not apply in the unsupervised continual learning dealing with the scarcity of labelled samples. The access of true class labels are only provided for the initial batch of each task to associate the cluster's centroids with classes $B_0^{k}=\{x_n,y_n\}_{n=1}^{N_{0}^k}$ while the remainder of data samples arrive with the absence of labelled samples $B_j^{k}=\{x_n\}_{n=1}^{N_j^k}$. Note that $N_0^k+N_1^k+N_j^k...+N_J^k=N_k$.

\section{Learning Policy of KIERA}
\subsection{Network Structure}
KIERA is structured as a SAE comprising two components: feature extraction layer and fully-connected layer. The feature extraction layer is built upon a convolutional layer or a nonlinear layer. It maps an input image $x$ into a latent input vector $Z\in\Re^{u'}$ where $u'$ denotes the number of natural features. The fully connected layer performs the encoding and encoding steps across $L_{sae}$ layers producing a reconstructed latent input vector $\hat{Z}$. The reconstructed latent input vector $\hat{Z}$ is further fed to transposed convolutional or nonlinear layers generating the reconstructed input image $\hat{x}$. 

Suppose that $h^{l-1}\in\Re^{u_l}$ stands for the latent input vector of the $l-th$ encoder where $h^{0}=Z$, the encoder projects $h^{l-1}$ to a lower dimensional latent space $h^{l}=r(W_{enc}^{l}h^{l-1}+b_l)$ while the decoder reconstructs the latent input vector $\hat{h}^{l-1}=r(W_{dec}^{l}+c_l)$. $r(.)$ is a ReLU activation function inducing a nonlinear mapping. $W_{enc}^{l}\in\Re^{R_l\times u_l},b_{l}\in\Re^{R_l}$ are the connective weight and bias of the $l-th$ encoder while $W_{dec}^{l}\in\Re^{u_l\times R_l},c_l\in\Re^{u_l}$ are the connective weight and bias of the $l-th$ decoder. $R_l,u_l$ are respectively the number of hidden nodes of the $l-th$ layer and the number of input features of the $l-th$ layer. The tied-weight constraint is applied here to avoid the over-training problem $W_{enc}^{l}=(W_{dec}^{l})^T$. The SAE carries out the non-linear dimension reduction step preventing the trivial solution as the case of the linear dimension reduction mechanism while the training process occurs in the greedy layer-wise fashion. 

The unique facet of KIERA lies in the different-depth network structure in which the clustering mechanism takes place in every hidden layer of the fully connected layer $h^l(.)$ thereby producing its own local output. It distinguishes itself from the conventional deep clustering network where the clustering mechanism happens at the bottleneck layer only. The cluster's allegiance \cite{smith2019unsupervised} is expressed:
\begin{equation}
    ale_{s,l}=\frac{\exp{-||C_{s,l}-h_{l}||}}{\max_{s=1,...,Cls_l}\exp{-||C_{s,l}-h_l||}}
\end{equation}
where $C_{s,l}$ denotes the centroid of the $s-th$ cluster of the $l-th$ layer and $Cls_l$ stands for the number of clusters of the $l-th$ layer. The cluster's allegiance is averaged across all labelled samples $B_0^k=\{x_i,y_i\}_{n=1}^{N_0^k}$ to indicate cluster's tendency to a target class. Let $N_0^{k,o}$ be the number of initially labelled samples of the $k-th$ task falling into the $o-th$ target class where $N_0^{k,1}+N_0^{k,o}+...+N_0^{k,m}=N_0^{k}$, the averaged cluster allegiance is written:
\begin{equation}\label{Ale}
    Ale_{s,l}^{o}=\frac{\sum_{n=1}^{N_{0}^{k,o}}ale_{s,l}^{n,o}}{N_{0}^{k,o}}
\end{equation}
\eqref{Ale} implies a low cluster allegiance of unclean clusters being populated by data samples of mixed classes. The score of the $o-th$ target class of the $l-th$ layer is computed by combining the cluster's allegiance and the distance to a data sample $h^{l}$:
\begin{equation}\label{Score}
    Score_{o}^{l}=Softmax(\sum_{s=1}^{Cls_l}\exp{-||h_l-C_{s,l}||}Ale_{s,l}^{o})
\end{equation}
The local output of the $l-th$ layer can be found by taking a maximum operation $\max_{o=1,...,m}Score_{o}^{l}$. The use of $Softmax(.)$ operation assures the partition of unity as well as the uniform range of $Score_{o}^{l}$ across $L_{sae}$ layers. The final output is produced by aggregating the local outputs using the summation operation as follows:
\begin{equation}
    \hat{y}=\max_{o=1,...,m}\sum_{l=1}^{L_{sae}}Score_{o}^{l}
\end{equation}
Labelled samples of each task $B_{0}^{k}$ are only exploited to associate a cluster to a specific class, i.e., the calculation of cluster allegiance \eqref{Ale} thereby satisfying the unsupervised learning requirement. Another typical characteristic of KIERA exists in its self-evolving property where $R_l,L_{sae},Cls_{l}$ are not deterministic rather self-evolved from data streams.   
\subsection{Structural Learning Mechanism}
\noindent\textbf{Evolution of Hidden Nodes:} KIERA features an elastic network width where new nodes are dynamically added while inactive nodes are pruned. This strategy is underpinned by the network significance (NS) method \cite{pratama2019automatic} estimating the generalization power of a network based on the bias and variance decomposition approach. New nodes are introduced in the case of high bias to cope with the underfitting issue while inconsequential nodes are discarded in the case of high variance to overcome the overfitting situation. Because of the absence of any labelled samples for model's updates, the network bias and variance are estimated from reconstructions losses of a layer $NS=Bias^{2}+Var=(E[\hat{h}_{l}]-h_{l})^2+(E[\hat{h}_l^2]-E[\hat{h}_l]^{2})$. Note that KIERA adopts the greedy layer-wise training approach. The network bias and variance are formalized under a normal distribution $p(x)=N(\mu,\sigma^{2})$ and the RelU activation function $r(.)$ where $\mu,\sigma^2$ are the mean and standard deviation of data samples. The modified statistical process control (SPC) method \cite{GamaDataStream}, a popular approach for anomaly detection, is utilized to signal the high bias condition and the high variance condition as follows:
	\begin{equation}\label{bias}
	    \mu_{bias}^n+\sigma_{bias}^n\geq\mu_{bias}^{min}+(\ln(l) + 1) \times k_1\sigma_{bias}^{min}
	\end{equation}
	\begin{equation}\label{variance}
	    \mu_{var}^n+\sigma_{var}^n\geq\mu_{var}^{min}+2 \times (\ln(l) + 1) \times k_2\sigma_{var}^{min}
	\end{equation}
where the main modification of the conventional SPC lies in $k_1=1.3\exp{(-Bias^2)}+0.7$ and $k_2=1.3\exp{(-Var^2)}+0.7$ leading to dynamic confidence levels. This idea enables new nodes to be crafted in the case of high bias and inconsequential nodes to be removed in the case of high variance. The Xavier's initialization approach is applied for the sake of initialization. Conversely, the pruning process targets an inactive node having the least statistical contribution $\min_{i=1,...,{R_l}}E[\hat{h}_{i}^{l}]$. The term $(\ln{(l)}+1)$ is inserted to guarantee the nonlinear feature reduction step due to $R_l<R_{l-1}$. The node growing strategy becomes insensitive as the increase of network depth. The term $2$ is applied in \eqref{variance} to prevent a new node to be directly eliminated. 
\newline\noindent\textbf{Evolution of Hidden Layer:} KIERA characterizes a variable network depth where a drift detection technique is applied to expand the network depth $L_{sae}$. A new layer is incorporated if a drift is detected. Note that insertion of a new layer is capable of boosting the network's capacity more significantly than that introduction of new nodes. Since no labelled samples are not available at all, the drift detection mechanism focuses on the dynamic of latent input features $Z$ extracted by the feature extraction layer. Nevertheless, the characteristic of latent input features are insensitive to changing data distributions. We apply the drift detection method for the last two consecutive data batches $A=[Z_{J-1}^{k},Z_{J}^{k}]$ here. The drift detection mechanism first finds a cutting point $cut$ signifying the increase of population means $\hat{A}+\epsilon_{A}\leq \hat{B}+\epsilon_{B}$. $\hat{A},\hat{B}$ are the statistics of two data matrices $A\in\Re^{2N_{j}^{k}\times u'}$ and $B\in\Re^{cut\times u'}$ respectively while $\epsilon_{A,B}=\sqrt{\frac{1}{2\times size}\ln{\frac{1}{\alpha}}}$ is their corresponding Hoeffding's bounds    and $\alpha$ is the significance level. $size$ is the size of data matrix of interest. 

Once finding the cutting point, the data matrix $A$ is divided into two matrices $B$ and $C^{(2N_{j}^k-cut)\times u'}$. A drift is confirmed if $|B-C|\geq\epsilon_{d}$ whereas a warning is flagged by $\epsilon_{d}\geq|B-C|\geq\epsilon_{w}$. $\epsilon_{d,w}=(b-a)\sqrt{\frac{(size-cut)}{2*size*cut}\ln{\frac{1}{\alpha_{d,w}}}}$ where $[a,b]$ represents the interval of the data matrix $A$ and $\alpha_d<\alpha_w$. The warning condition is a transition situation where the drift condition is to be confirmed by a next stream. A new layer is added if a drift condition is signalled. The number of nodes of the new layer is set at the half of that of the previous layer. Addition of a new layer does not impose the catastrophic forgetting because of the different-depth network structure, i.e., every layer to produce its local output.  
\newline\noindent\textbf{Evolution of Hidden Clusters:} the growing strategy of hidden clusters relies on the compatibility measure examining a distance between a latent sample $h^l$ to the nearest cluster. A new cluster is added if a data sample is deemed remote to the zone of influence of any existing clusters as follows:
\begin{equation}\label{clust_update}
    \min_{s=1,...,Cls_l}D(C_{s,l},h^l)>\mu_{D_{s,l}}+k_3*\sigma_{D_{s,l}}
\end{equation}
where $\mu_{D_{s,l}},\sigma_{D_{s,l}}$ are the mean and standard deviation of the distance measure $D(C_{s,l},h^l)$ while $k_3=2\exp{-||h^l-C_{s,l}||}+2$. \eqref{clust_update} is perceived as the SPC method with the dynamic confidence degree assuring that a cluster is added if it is far from the coverage of existing clusters. \eqref{clust_update} hints the presence of a new concept unseen in the previous observations. Hence, a data sample $h^l$ can be regarded as a focal point. A new cluster is integrated by setting the current sample as a centroid of a new cluster $C_{(s+1),l}=h^l$ while its cardinality is set as $Car_{(s+1),l}=1$. Note that the clustering process occurs in different levels of deep embedding space.
\subsection{Parameter Learning Mechanism}
\noindent\textbf{Network Parameters:} the parameter learning strategy of network parameters performs simultaneous feature learning and clustering in which the main goal is to establish clustering friendly latent spaces \cite{clustering_friendly}. The loss function comprises two terms: reconstruction loss and clustering loss written as follows:
\begin{equation}\label{loss}
    L_{all}=\underbrace{L(x,\hat{x})}_{L_1}+\sum_{l=1}^{L_{sae}}\underbrace{(L(h^l,\hat{h}^l)+\frac{\lambda}{2}||h^{l}-C_{win,l}||_2)}_{L_2}
\end{equation}
where $\lambda$ is a tradeoff constant controlling the influence of each loss function. \eqref{loss} can be solved using the stochastic gradient descent optimizer where $L_1$ is executed in the end-to-end fashion while $L_2$ is carried out per layer, i.e., greedy layer wise fashion. $C_{win,l}$ stands for the centroid of the winning cluster of the $l-th$ layer, i.e., the closest cluster to a latent sample $win\rightarrow \min_{s=1,...,Cls_l}||h^l-C_{s,l}||$. The first and second terms $L(x,\hat{x}),L(h^l,\hat{h}^l)$ are formed as the mean squared error (MSE) loss function where $L(x,\hat{x})$ assures data samples to be mapped back to their original representations while $L(h^l,\hat{h}^l)$ is to extract meaningful latent features in every hidden layer of SAE. The last term $||h^l-C_{win,l}||_2$ is known as the distance loss or the K-means loss producing the K-means friendly latent space. That is, a latent sample is driven to be close to the centroid of the winning hidden cluster resulting in a high cluster probability, i.e., the assignment probability of a data sample to its nearest cluster is high. The multiple nonlinear mapping via the SAE also functions as nonlinear feature reduction addressing the trivial solution as often the case of linear mapping. 
\newline\noindent\textbf{Cluster Parameters:} the parameter learning of hidden clusters is executed if \eqref{clust_update} is violated meaning that a latent sample $h^l$ is sufficiently adjacent to existing clusters. This condition only calls for association of the current latent sample to the winning cluster, the nearest cluster, fine-tuning the centroid of the winning cluster as follows:
\begin{equation}\label{clust_update1}
    C_{win,l}=C_{win,l}-\frac{(C_{win,l}-h^l)}{Car_{win,l}+1}; Car_{win,l}=Car_{win,l}+1
\end{equation}
The tuning process improves the coverage of the winning cluster to the current sample and thus enhances the cluster's posterior probability $\uparrow P(C_{win,l}|x)$. The intensity of the tuning process in (\eqref{clust_update1}) decreases as the increase of cluster's cardinality thereby expecting convergence. Only the winning cluster is adjusted here to avoid the issue of cluster's overlapping. The alternate optimization strategy is implemented in KIERA where the cluster's parameters are fixed while adjusting the network parameters and vice versa.
\subsection{Centroid-based Experience Replay}
KIERA adopts the centroid-based experience replay strategy to address the catastrophic forgetting problem. This mechanism stores focal-points of previous tasks $T_1,T_2,...,T_{k-1}$ into an episodic memory interleaved along with samples of the current task. $T_t$ to overcome the catastrophic interference issue. Note that unlabelled images are retained in the episodic memory. The sample selection mechanism is driven by the cluster growing strategy in \eqref{clust_update}. That is, an image is considered as a focal-point $x^*$ thus being stored in the episodic memory $M_{em}^{k}=M_{em}^{k-1}\cup x^*$ provided that \eqref{clust_update} is observed. The sample selection strategy is necessary to control the size of memory as well as to substantiate the efficacy of experience replay mechanism making sure conservation of important images, focal points. Focal points represent varieties of concepts seen thus far.  

The size of memory grows as the increase of tasks making the experience replay mechanism intractable for a large problem. On the other hand, the network parameters and the cluster parameters are adjusted with recent samples thereby paving possibility of the catastrophic forgetting issue. A selective sampling strategy is implemented in the centroid-based experience replay method. The goal of the selective sampling strategy is to identify the most forgotten focal points in the episodic memory for the sake of experience replay while ignoring other focal-points thereby expediting the model's updates. The most forgotten samples are those focal points which do not receive sufficient coverage of existing clusters. That is, the cluster posterior probabilities $P(C_{s,l}|x^*)=\exp{-||C_{s,l}-h_{*}^l||}$ of the most forgotten samples are below a midpoint $mid$ determined:
\begin{equation}
    mid=\frac{\sum_{n=1}^{N_{em}}\max_{s=1,...,Cls_l}P(C_{s_l}|x^*)}{N_{em}}
\end{equation}
where $N_{em}$ denotes the size of episodic memory. The midpoint indicates the average level of coverage to all focal-points in the episodic memory. A focal-point is included into the replay buffer $B^*=B^*\cup x^*$ to be replayed along with the current concept if it is not sufficiently covered $P(C_{s,l}|x^*)< mid$. Finally, the centroid-based experience replay strategy is executed by interleaving images of current data batch $B^k$ and replay buffer $B^*$ for the training process $B^k\cup B^*$. The most forgotten focal points are evaluated by checking the current situation of network parameters and cluster parameters portrayed by the cluster posterior probability. 

\section{Proof of Concepts}
\subsection{Datasets}
The performance of KIERA is numerically validated using four popular continual learning problems: Permutted MNIST (PMNIST), Rotated MNIST (RMNIST), Split MNIST (SMNIST) and Split CIFAR10 (SCIFAR10). PMNIST is constructed by applying four random permutations to the original MNIST problem thus leading to four tasks $K=4$. RMNIST applies rotations with random angles to the original MNIST problem: $[0,30],[31,60],[61,90],[91,120]$ thus creating four tasks $K=4$ in total. The two problems characterize the concept drift problem in each task. The SMNIST presents the incremental class problem of five tasks where each task presents two mutually exclusive classes,$(0/1,2/3,4/5,6/7,8/9)$. As with the SMINST problem, the SCIFAR10 also features the incremental class problem of five tasks where each task possesses two non-overlapping target classes. 
\subsection{Implementation Notes of KIERA}
KIERA applies an iterative training strategy for the initial training process of each task and when a new layer is added. This mechanism utilizes $N_{init}^k$ unlabelled samples to be iterated across $E$ number of epochs where $N_{init}^k$ and $E$ are respective selected as 1000 and 50 respectively.  The training process completely runs in the single-pass training mode afterward. $B_0^{k}=\{x_i,y_i\}_{i=1}^{N_{0}^{k}}$ labelled samples are revealed for each task to associate a cluster with a target class in which for every class $100$ labelled samples are offered. Hence, $N_0^k$ is 200 for the SMNIST problem and the SCIFAR10 while $N_0^k$ is 1000 for RMNIST and PMNIST.  
\subsection{Network Structure}
The feature extraction layer of KIERA is realized as the convolutional neural network (CNN) where the encoder part utilizes two convolutional layers with 16 and 4 filters respectively and max-pooling layer in between. The decoder part applies transposed convolutional layers with 4 and 16 filters respectively. For PMNIST problem, the feature extraction layer is formed as a multilayer perceptron (MLP) network with two hidden layers where the number of nodes are selected as $[1000,500]$. Note that the random permutation of PMNIST requires a model to consider all image pixels as done in MLP. The fully connected layer is initialized as a single hidden layer $L_{sae}=1$ with $96$ hidden nodes $R_1=96$. The ReLU activation function is applied as the hidden nodes and the sigmoid function is implemented in the decoder output to produce normalized reconstructed images.

\subsection{Hyper-parameters}
The hyper-parameters of KIERA are fixed across the four problems to demonstrate non ad-hoc characteristic. The learning rate, momentum coefficient and weight decay strength of the SGD method are selected as $0.01,0.95,5\times 10^{-5}$ while the significant levels of the drift detector are set as $\alpha=0.001,\alpha_d=0.001,\alpha_w=0.005$. The trade-off constant is chosen as $\lambda=0.01$. 
\subsection{Baseline Algorithms}
KIERA is compared against Deep Clustering Networks (DCN) \cite{clustering_friendly}, AE+KMeans and STAM \cite{smith2019unsupervised}. DCN adopts simultaneous feature learning and clustering where the cost function is akin to KIERA \eqref{loss}. Nevertheless, it adopts a static network structure. AE+KMeans performs the feature learning first using the reconstruction loss while the KMeans clustering process is carried out afterward. STAM \cite{smith2019unsupervised} relies on an irregular feature extraction layer based on the concept of patches while having a self-clustering mechanism as with KIERA. DCN and AE+KMeans are fitted with the Learning Without Forgetting (LWF) method \cite{li2016learning} and Synaptic Intelligence (SI) method \cite{zenke2017continual} to overcome the catastrophic forgetting problem. DCN and AE+KMeans make use of the same network structure as KIERA to ensure fair comparison. The regularization strength of LWF. is set as $\beta=5$ while it is allocated as $0.2$ for the first task and $0.8$ for the remaining task in SI method. 

The hyper-parameters of STAM are selected as per their original values but hand-tuned if its performance is surprisingly poor. The baseline algorithms are executed in the same computational environments using their published codes. The performance of consolidated algorithms are examined using four evaluation metrics: prequential accuracy (Preq Acc), task accuracy (Task Acc), backward transfer (BWT) and forward transfer (FWT). Preq Acc measures the classification performance of the current task while Task Acc evaluates the classification performance of all tasks after completing the learning process. BWT and FWT are put forward in \cite{gem2017paz} where FWT indicates knowledge transfer across task while BWT reveals knowledge retention of a model after learning a new task. BWT and FWT ranges in $-\infty$ and $+\infty$ with a positive high value being the best value. All consolidated algorithms are run five times where the averages are reported in Table \ref{unsupervisedCL}. 
\subsection{Numerical Results}
Referring to Table \ref{unsupervisedCL}, KIERA delivers the highest Preq accuracy and FWT in the rotated MNIST problem with statistically significant margin while being on par with STAM in the context of Task Acc.  KIERA outperforms other algorithms in the PMNIST problem in which it obtains the highest Task Acc and Preq Acc with substantial differences to its competitors. Although the FWT of DCN+LWF is higher than KIERA, its Task Acc and its Preq Acc are poor. Similar finding is observed in the SCIFAR10 problem, where KIERA outperforms other algorithms in the Preq Acc, Task Acc and FWT with statistically significant differences. KIERA does not perform well only on the SMNIST problem but it is still much better than DCN and AE+KM. 

Despite its competitive performance, STAM takes advantage of a non-parametric feature extraction layer making it robust against the issue of catastrophic forgetting. This module, however, hinders its execution under the GPU environments thus slowing down its execution time. The advantage of clustering-based approach is seen in the FWT aspect. Although STAM adopts the clustering approach, its network parameters are trained with the absence of clustering loss. The self-evolving network structure plays vital role here where Task Acc and Preq Acc of KIERA and STAM outperforms DCN and AE+KMeans having fixed structure in all cases with noticeable margins. The centroid-based experience replay mechanism performs well compared to SI and LWF. 
\begin{table}[t!]
\caption{Numerical Results of Consolidated Algorithms.}
\label{unsupervisedCL}
\begin{center}
\scalebox{0.62}{
\begin{tabular}{llcccc}
\toprule
Datasets & Methods & BWT & FWT & Task Acc. (\%) & Preq. Acc. (\%) \\
\midrule
RMNIST & KIERA & -7 $\pm$ 1.53 & \textbf{44 $\pm$ 1.08} & {76.84 $\pm$ 0.53} & \textbf{79.91  $\pm$ 0.47} \\
 & STAM & \textbf{0.9 $\pm$ 0.35} & 30 $\pm$ 0.37$^\times$ & \textbf{77.58 $\pm$ 0.43} & 74.71 $\pm$ 0.29$^\times$ \\
 & DCN+LwF & -15 $\pm$ 6.54$^\times$ & 16 $\pm$ 5.50$^\times$ & 39.47 $\pm$ 10.13$^\times$ & 52.79 $\pm$ 13.54$^\times$ \\
 & DCN+SI & -13 $\pm$ 6.00$^\times$ & 17 $\pm$ 6.92$^\times$ & 44.62 $\pm$ 12.46$^\times$ & 55.66 $\pm$ 14.83$^\times$ \\
 & AE+KM+LwF & -18 $\pm$ 2.32$^\times$ & 18 $\pm$ 1.79$^\times$ & 45.31 $\pm$ 1.63$^\times$ & 60.15 $\pm$ 1.54$^\times$\\
 & AE+KM+SI & -9 $\pm$ 2.72$^\times$ & 16 $\pm$ 2.15$^\times$ & 49.07 $\pm$ 0.73$^\times$ & 51.19 $\pm$ 1.39$^\times$\\
 \midrule
PMNIST & KIERA & -22 $\pm$ 3.22 & 2 $\pm$ 0.77 & \textbf{59.90 $\pm$ 2.48} & \textbf{74.59 $\pm$ 0.5} \\
& STAM & \textbf{0.3 $\pm$ 0.09} & {1 $\pm$ 0.44}$^\times$ & 47.97 $\pm$ 0.59$^\times$ & 55.37 $\pm$ 0.26$^\times$ \\
 & DCN+LwF & -30 $\pm$ 1.70$^\times$ & \textbf{3  $\pm$ 1.42} & {35.53  $\pm$ 0.78}$^\times$ & 56.50 $\pm$ 0.54$^\times$ \\
 & DCN+SI & -43 $\pm$ 3.23$^\times$ & 1 $\pm$ 1.01$^\times$ & 33.09  $\pm$ 2.14$^\times$ & {64.87 $\pm$ 0.31}$^\times$ \\
 & AE+KM+LwF & -28 $\pm$ 1.71$^\times$ & {3  $\pm$ 1.85} & 35.53  $\pm$ 1.02$^\times$ & 56.27 $\pm$ 0.55$^\times$ \\
 & AE+KM+SI & -35 $\pm$ 2.35$^\times$ & 1 $\pm$ 1.97$^\times$ & 36.06 $\pm$ 1.23$^\times$ & 61.50 $\pm$ 0.36$^\times$ \\
\midrule
 SMNIST & KIERA & -9 $\pm$ 1.15 & 15 $\pm$ 3.56 & {84.29 $\pm$ 0.92} & {91.06 $\pm$ 0.71} \\
 & STAM & \textbf{-2 $\pm$ 0.13} & 0 & \textbf{92.18 $\pm$ 0.32} & \textbf{91.98 $\pm$ 0.32} \\
 & DCN+LwF & -7 $\pm$ 3.98 & 18 $\pm$ 1.17 & 52.42 $\pm$ 5.02$^\times$ & 53.46 $\pm$ 2.25$^\times$ \\
 & DCN+SI & -4 $\pm$ 1.45 & \textbf{22 $\pm$ 2.89} & 58.82 $\pm$ 1.18$^\times$ & 57.00 $\pm$ 0.34$^\times$ \\
 & AE+KM+LwF & -5 $\pm$ 1.11 & 18 $\pm$ 1.03 & 55.12 $\pm$ 0.97$^\times$ & 54.69 $\pm$ 0.58$^\times$ \\
 & AE+KM+SI & {-3 $\pm$ 1.88} & 22 $\pm$ 1.73 & 58.84 $\pm$ 0.71$^\times$ & 56.58 $\pm$ 0.28$^\times$\\
 \midrule
 SCIFAR10 & KIERA & -15 $\pm$ 5.92 & \textbf{15 $\pm$ 1.30} & \textbf{25.64 $\pm$ 1.85} & \textbf{37.09 $\pm$ 2.83} \\
 & STAM & -18 $\pm$ 2.46 & 0$^\times$ & 20.60 $\pm$ 0.66$^\times$ & 35.43 $\pm$ 1.22 \\
 & DCN+LwF & -15 $\pm$ 1.32 & 6 $\pm$ 1.67$^\times$ & 14.87 $\pm$ 1.66$^\times$ & 23.98 $\pm$ 1.17$^\times$ \\
 & DCN+SI & {-16 $\pm$ 1.68} & 6 $\pm$ 1.90$^\times$ & 17.15 $\pm$ 1.14$^\times$ & 25.51 $\pm$ 0.50$^\times$ \\
 & AE+KM+LwF & \textbf{-12 $\pm$ 2.00} & 8 $\pm$ 0.63$^\times$ & 22.12  $\pm$ 0.33$^\times$ & 28.77 $\pm$ 0.84$^\times$ \\
 & AE+KM+SI & -13 $\pm$ 1.36 & 9 $\pm$ 1.34$^\times$ & 21.91 $\pm$ 0.68$^\times$ & 28.95 $\pm$ 1.11$^\times$ \\
\bottomrule
\end{tabular}
}
\end{center}
\footnotesize{$^{\times}$: Indicates that the numerical results of the respected baseline and KIERA are significantly different.}
\end{table}

Table \ref{finalState} reports the network structures of KIERA and the episodic memory. The structural learning of KIERA generates a compact and bounded network structure where the number of hidden nodes, hidden layer and hidden clusters are much less than the number of data points. The hidden layer evolution is seen in the PMNIST problem where additional layers are inserted. The centroid-based experience replay excludes $N_{init}^k$ unlabelled samples for the pre-training phase of each task. As a result, the focalpoints of the episodic memory are less than those the number of clusters.  
\begin{table}[t!]
\caption{The Final State of KIERA.}
\label{finalState}
\begin{center}
\scalebox{0.8}{
\begin{tabular}{lcccc}
\toprule
Datasets &  NoN & NoL & NoC (K) & NoM (K) \\
\midrule
RMNIST &  101 $\pm$ 1 & {1 $\pm$ 0} & 3.2 $\pm$ 0.02 & 1.2 $\pm$ 0.05 \\
PMNIST &  162 $\pm$ 12 & 3 $\pm$ 1 & 5.4 $\pm$ 0.64 & 0.8 $\pm$ 0.15 \\
SMNIST &  95 $\pm$ 3 & 1 $\pm$ 0 & 2.8 $\pm$ 0.13 & 1 $\pm$ 0.1 \\
SCIFAR10 & (1.5 $\pm$ 2.8)K & 1 $\pm$ 0 & 4.2 $\pm$ 0.44 & 2.3 $\pm$ 0.16 \\
\bottomrule
\end{tabular}
}
\end{center}
\footnotesize{NoN: total number of hidden nodes; NoL: total number of hidden layers; NoC: total number of hidden clusters; NoM: number of samples in episodic memory.}
\end{table}

\begin{figure}[!t]
\centerline{\includegraphics[scale=0.5]{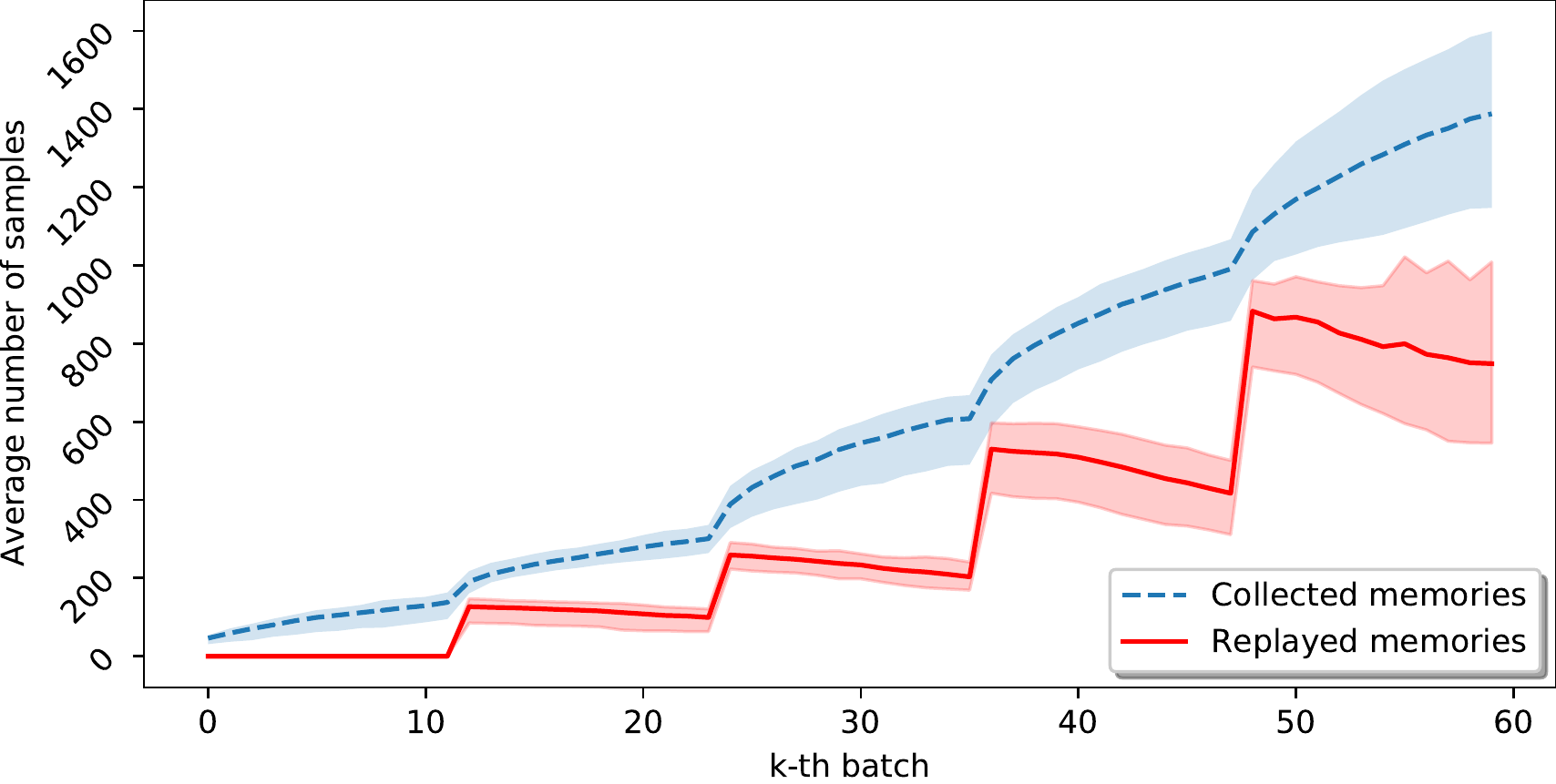}}
\caption{The Comparison of Average Collected and Replayed Memories on SMNIST Problem.}
\label{fig:memory}
\end{figure}

Fig. \ref{fig:memory} exhibits the evolution of episodic memory $M_{em}$ and replay buffer $B^*$. It is shown that the episodic memory $M_{em}$ grows in much faster rate than the replay buffer $B^*$. This trend becomes obvious as the increase of the tasks. This finding is reasonable because each task possesses concept changes inducing addition of new clusters and thus focal-points of the episodic memory. The selective sampling is capable of finding the most forgotten samples thus leading to bounded size of the replay buffer. The number of forgotten samples is high in the beginning of each task but reduces as the execution of centroid-based experience replay.  

\section{Conclusion}
This paper presents an unsupervised continual learning approach termed KIERA built upon the flexible clustering principle. KIERA features the self-organizing network structure adapting quickly to concept changes. The centroid-based experience replay is proposed to address the catastrophic forgetting problem. Our numerical study with four popular continual learning problems confirm the efficacy of KIERA in attaining high Preq Acc, high Task Acc and high FWT. KIERA delivers higher BWT than those using SI and LWF. The advantage of structural learning mechanism is also demonstrated in our numerical study where it produces significantly better performance than those static network structure. KIERA is capable of learning and predicting without the presence of Task ID and Task's boundaries. Our memory analysis deduces the effectiveness of the centroid-based experience replay in which the size of replay buffer is bounded. Our future study answers the issue of multiple streams.  
\bibliographystyle{named}
\bibliography{ijcai21}
\end{document}